\documentclass[pmlr]{jmlr}

\usepackage{multirow}
\usepackage{mathtools}
\DeclarePairedDelimiter{\norm}{\lVert}{\rVert}
\usepackage{caption}
\usepackage{booktabs}
\usepackage[english]{babel}

\usepackage{longtable}
\usepackage{bbm}
\makeatletter
\let\Ginclude@graphics\@org@Ginclude@graphics 
\makeatother

\jmlrvolume{260}
\jmlryear{2024}
\jmlrworkshop{ACML 2024}
\jmlrpages{-}
\renewcommand{\thepage}{}

\title[AP-LSE]{Robust Transfer Learning for Active Level Set Estimation with Locally Adaptive Gaussian Process Prior}
  \author{\Name{Giang Ngo\nametag{\thanks{corresponding author}}} \Email{g.ngo@deakin.edu.au}\\
   \Name{Dang Nguyen} \Email{d.nguyen@deakin.edu.au}\\
   \Name{Sunil Gupta} \Email{sunil.gupta@deakin.edu.au}\\
   \addr Applied Artificial Intelligence Institute (A2I2), Geelong, Victoria, Australia}
\editors{Vu Nguyen and Hsuan-Tien Lin}

\begin{document}

\maketitle

\begin{abstract}
The objective of active level set estimation for a black-box function is to precisely identify regions where the function values exceed or fall below a specified threshold by iteratively performing function evaluations to gather more information about the function. This becomes particularly important when function evaluations are costly, drastically limiting our ability to acquire large datasets. A promising way to sample-efficiently model the black-box function is by incorporating prior knowledge from a related function. However, this approach risks slowing down the estimation task if the prior knowledge is irrelevant or misleading. In this paper, we present a novel transfer learning method for active level set estimation that safely integrates a given prior knowledge while constantly adjusting it to guarantee a robust performance of a level set estimation algorithm even when the prior knowledge is irrelevant. We theoretically analyze this algorithm to show that it has a better level set convergence compared to standard transfer learning approaches that do not make any adjustment to the prior. Additionally, extensive experiments across multiple datasets confirm the effectiveness of our method when applied to various different level set estimation algorithms as well as different transfer learning scenarios.
\end{abstract}

\section{Introduction}
The level set estimation (LSE) problem involves classifying whether \(\textbf{x}\) belongs to the superlevel set (i.e. \(f(\textbf{x})\) exceeds a threshold \(h\)) or the sublevel set (i.e. \(f(\textbf{x})\) is below \(h\)). LSE arises in many important applications including environmental monitoring (\cite{activelse}), computational chemistry (\cite{10.1063/5.0044989}), and material characterization (\cite{hozumi2023adaptive}). Two common difficulties of these applications are that the underlying function relating its input variables to the output is unknown, i.e. a \textit{black-box} function, and that evaluating such black-box functions requires conducting real-world experiments that are often very expensive. To model the black-box function for level set classification in such scenarios, an intuitive approach is to use an active learning method to sequentially acquire new data points that are highly informative for determining the level set. The focus of active LSE algorithms is often on finding the locations of those informative data points at each iteration. This is done by optimizing an \textit{acquisition function}, which is an indicator of how useful a data point is in classifying the level sets. The acquisition function is constructed using a surrogate model of the black-box function built with the existing function evaluations. The black-box function is then evaluated at the next point suggested by the acquisition function, and this new observation is used to update the surrogate model. The surrogate model will be used to classify the level set of any point either at every iteration or when an evaluation budget runs out.

Due to the high cost of function evaluations, leveraging additional relevant information can significantly reduce the number of evaluations required to achieve the desired estimation accuracy. Such relevant information may come from prior information based on domain knowledge or data from a related LSE task completed in the past. For instance, in the context of \textit{algorithmic assurance} (\cite{NEURIPS2018_cc709032}), where the objective is to identify scenarios in which the performance of a pretrained machine learning model reaches a certain threshold, prior information can be sourced from the identified scenarios of a related model. Other examples include the elicitation of expert knowledge to form a prior function to model the output of complex computer code by \cite{10.1111/1467-9884.00300} or the problem of discovering Higgs boson when \cite{Golchi} suggested the construction of a prior mean function by either Monte Carlo studies or a polynomial fit to simulated data.

The previous works on using prior information for active LSE are limited. Only \cite{hozumi2023adaptive} attempted the problem but assumed that data from a related task were available. There have been plenty of attempts to use prior information in a related problem known as Bayesian Optimization (BO), where the prior information has been used to accelerate the finding of the optimum of a function (\cite{NIPS2013_f33ba15e, diffgp, wistuba2015hyperparameter}). However, since the function at hand is black-box, assessing the relevance of prior information is not always straightforward. At times, it can be misleading and hinder the LSE task. Though the prior information is believed to be related to \(f\), mismatches can occur, either entirely or partially across the input space. For example, in environmental monitoring, the condition of the whole environment or part of it can change compared to the previous LSE task. Existing LSE algorithms often employ Gaussian Process (GP) (\cite{rasmussen}) as the surrogate model in which the posterior mean prediction at \(\textbf{x}\) can combine information from existing data points and the prior distribution. In case of mismatches, the posterior mean that uses prior information will be less accurate than the one without it.

This inaccuracy in modeling \(f\) will weaken the sample efficiency and accuracy of an LSE algorithm. Firstly, it will be harder for the acquisition function to correctly indicate how informative a point is in level set classification. One popular criterion for acquisition functions is to prioritize data points that are near the threshold boundary (\cite{straddle, rmile, activelse}), and this becomes non-trivial if the acquisition function fails to tell if the function value at a point is close to \(h\) or not due to inaccurate function estimation. The algorithm then needs to explore more to locate the targeted points. Another obvious consequence is inaccurate level set classification where the function is not estimated accurately, lowering the accuracy of the LSE algorithm. The GP posterior will need more data points to mitigate the impact of inaccurate prior information in a region, which again hurts the sample efficiency of the algorithm.

Given the above issues, it is necessary to control the use of prior information so that it will not adversely affect the LSE algorithm in regions where it is observed to be irrelevant to \(f\). However, we would still like to use the prior information in regions with insufficient data points. This paper proposes a \textit{locally adaptive prior} that adjusts the prior function according to the mismatches between itself and the observed function evaluations. The adjusted prior becomes similar to the observed function values in the discovered regions and reverts to its own form in undiscovered regions, allowing a robust transfer of prior knowledge into an LSE algorithm regardless of any potential unseen mismatch between the prior knowledge and the actual underlying function. Thus, the prior information is safely transferred into the surrogate model without letting the observed mismatches affect the main LSE algorithm. 

Our main contributions are:
\begin{enumerate}
    \item A novel method to safely transfer prior information in active LSE to accelerate the estimation of level sets for a black-box function.
    \item A theoretical analysis of the convergence of our method as well as its efficacy compared to transfer learning using priors without any adaptations.
    \item Extensive empirical evidence showing robust performance of our method on multiple LSE tasks, on different LSE algorithms, with different levels of mismatches.
\end{enumerate}

\section{Related Works}
\textbf{Level Set Estimation}

\noindent The main difference among LSE algorithms is in the selection of a new data point to evaluate its function value at each iteration. The selected point needs to reveal as much information about the level sets of \(f\) as possible. A popular direction in active LSE is to select the point where the surrogate model is the most uncertain in performing level set classification. \cite{straddle} proposed maximizing the Straddle heuristic \(1.96\sigma_t(\textbf{x})-|\mu_t(\textbf{x})-h|\) where \(\mu_t(\textbf{x})\) and \(\sigma_t(\textbf{x})\) are the posterior mean and standard deviation of a GP surrogate model for \(f\) at iteration \(t\). This acquisition function prioritizes a point either in the undiscovered region or near the threshold boundary. ActiveLSE (\cite{activelse}) generalized the Straddle heuristic by maximizing the classification ambiguity \(\textup{min}\{\textup{max}(C_t(\textbf{x})-h,h-\textup{min}(C_t(\textbf{x})))\}\) in which \(C_t(\textbf{x})\) is the successive intersection of the confidence intervals of \(f(\textbf{x})\) at each iteration. Due to the maintenance of \(C_t(\textbf{x})\), ActiveLSE needs to operate on a fixed discrete number of data points while the Straddle heuristic can work on a continuous domain, and the two algorithms often share similar empirical performance. C2LSE (\cite{c2lse}) attempted to select points with the smallest classification confidence by maximizing \(\sigma_t(\textbf{x})/\textup{max}(\epsilon,|\mu_t(\textbf{x})-h|)\) where \(\epsilon>0\) prevents the algorithm from getting stuck at the boundary and also controls the algorithm's accuracy.

Another direction of active LSE is related to one-step lookahead mechanism. TruVar (\cite{truvar}) chose a point such that it minimizes the sum of truncated variances over a set of fixed points that arise from choosing that point. This algorithm also considers point-wise evaluation costs to favor locations with cheaper evaluations. RMILE (\cite{rmile}) is another algorithm within this category that is influenced by the expected improvement algorithm. Its acquisition function calculates the expected change in volume of the superlevel set if a point \(\textbf{x}\) is chosen and also takes into account the uncertainty at a point to avoid getting stuck. Other approaches for active LSE include hierarchical space partitioning (\cite{multiscale}), information-theoretic acquisition (\cite{infotheory}), and using experimental design (\cite{exdesign}).

\noindent\textbf{Transfer Learning for Level Set Estimation}

\noindent There are limited works in transfer learning for active LSE. \cite{hozumi2023adaptive} used an active LSE approach to identify defective areas on silicon ingots for solar cell production where they utilized measurements on a previous ingot by employing Diff-GP (\cite{diffgp}), which is originally designed for BO to incorporate data from a source task into the target task by modeling the difference between the two tasks. BO is often seen as related to active LSE where its goal is to find the optimum of a black-box function, and BO also often employs a GP as its surrogate model. Existing works in transfer learning for BO include jointly learned kernel in multi-task setting (\cite{NIPS2013_f33ba15e,7822140}), seeing source data as noisy observation of the target task (\cite{diffgp, THECKELJOY2019656}), and search space design (\cite{wistuba2015hyperparameter, perrone2019learning}). These works also consider the setting where prior information is from available source data, so they are not applicable to the setting where prior information is from domain knowledge.
\section{Proposed Method}
We propose a novel method to integrate prior information into the active LSE process. Our main idea is to adapt the prior locally based on any observed discrepancies before using it for the active LSE.
\subsection{Problem Formulation}
We now formally define the active LSE problem when \textit{prior knowledge is available}.

\noindent\textbf{Active level set estimation problem}: The goal is to iteratively obtain new data points of a black-box function \(f: \mathcal{X}\rightarrow \mathbb{R}\) with compact set \(\mathcal{X}\in \mathbb{R}^d\) to improve a classifier which accurately determines whether any point \(\textbf{x}\in\mathcal{X}\) is in the superlevel set \(H=\{\textbf{x}\in \mathcal{X}|f(\textbf{x})> h\}\) or the sublevel set \(L=\{\textbf{x}\in \mathcal{X}|f(\textbf{x}) < h\}\) given a threshold \(h \in \mathbb{R}\).

\noindent\textbf{Using prior knowledge}: We assume that the prior knowledge about \(f\) is in the form of \(u_p: \mathcal{X}\rightarrow \mathbb{R}\). This prior knowledge may be available through a domain-specific simulator, approximate models, or as a result of fitting a machine learning model on the data from a related setting. Our goal is to leverage \(u_p\) in active LSE for \(f\) such that we either need fewer data points or achieve more accurate classification than starting active LSE from scratch.
\subsection{Prior in Gaussian Process}
GPs are often used as the surrogate models for \(f\) in the problem of active LSE. A GP is specified by its mean function \(\mu(\textbf{x})\) and its kernel \(k(\textbf{x}, \textbf{x}')\). The posterior distribution over \(f\) of a GP given its zero mean prior and noisy observations is as follows:
\begin{align}
    \mu_t(\textbf{x})&=\textbf{k}_t(\textbf{x})^T(\textbf{K}_t+\sigma^2\mathcal{I})^{-1}\textbf{Y}_t\label{mu}\\
    \sigma_t^2(\textbf{x})&=k(\textbf{x},\textbf{x})-\textbf{k}_t(\textbf{x})^T(\textbf{K}_t+\sigma^2\mathcal{I})^{-1}\textbf{k}_t(\textbf{x})
\end{align}
where \(\textbf{k}_t(\textbf{x})=[k(\textbf{x}, \textbf{x}_i)]_{i=1}^t\), \(\textbf{K}_t=[k(\textbf{x}_i,\textbf{x}_j)]_{i,j=1}^t\), and \(\textbf{Y}_t=f(\textbf{X}_t)+\textbf{E}_t\) are the noisy observations of \(f\) at \(\textbf{X}_t=[\textbf{x}_i]_{i=1}^t\) with observation noises \(\textbf{E}_t=[\eta_i]_{i=1}^t\), \(\eta_i\sim \mathcal{N}(0,\sigma^2)\).

A non-trivial prior function \(u_p\) can be incorporated in the posterior mean distribution of such Gaussian Processes by seeing it as the prior mean function as follows:
\begin{align}
    \Bar{\mu}_t(\textbf{x})&=u_p(\textbf{x})+\textbf{k}_t(\textbf{x})^T(\textbf{K}_t+\sigma^2\mathcal{I})^{-1}(\textbf{Y}_t-u_p(\textbf{X}_t))\label{mu_bar}
\end{align}

\begin{figure}
    \centering
    \includegraphics[width=0.4\textwidth]{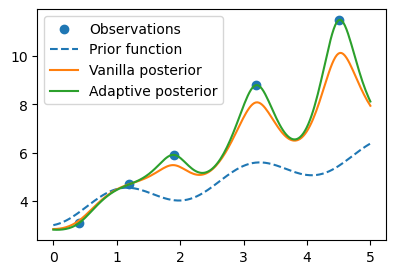}
    \caption{Blue: prior function; Orange: posterior mean by vanilla transfer learning; Green: posterior mean with our method AP-LSE. For the two data points on the left when the prior is still relevant compared to the observed function values, the posterior mean obtained with vanilla transfer learning still closely matches the observed values. When the difference is large as seen on the right, \(\Bar{\mu}_t\) starts to deviate from the data points. On the other hand, the posterior mean obtained with AP-LSE remains better fitted in both situations.}
    \label{fig:motivation}
    \vspace{-0.5cm}
\end{figure}

Directly using \(u_p\) as in Eq.\ref{mu_bar}, which we refer to as \emph{vanilla transfer learning} for GP, can help to transfer possibly related knowledge into the posterior mean of the Gaussian Process. However, it is possible that with the observations of \(f\), one may discover differences between \(f\) and \(u_p\) at certain points in \(\textbf{X}_t\), leading to inaccurate estimation of \(f\) due to the bias introduced by \(u_p\) as seen in Figure \ref{fig:motivation}. As explained in Section 1, mitigating the effect of \(u_p\) in the posterior mean around these points is necessary for active LSE tasks.
\subsection{Locally adaptive prior for active level set estimation}
The challenge is in adjusting \(u_p\) around observed mismatching locations but still transferring prior knowledge into the GP at other locations where we have no data. A simple solution is to define a region in \(\mathcal{X}\) that surrounds the observed data points where the prior function \(u_p\) is not used, and the posterior mean becomes \(\mu_t\) as in Eq.\ref{mu} instead. However, such a region will disrupt the smoothness of the posterior mean and likely make acquisition optimization difficult. It is also challenging to properly define the region since there is little information to determine the effective radius (i.e. how far away from \(\textbf{X}_t\) is sufficient).

To address this challenge, we propose \textbf{Adaptive Prior Level Set Estimation} (AP-LSE) which smoothly adjusts the prior function \(u_p\) by an additive term \(u_{dt}\) according to the observed difference between the current observations and the prior function, i.e. \(\textbf{Y}_t-u_p(\textbf{X}_t)\), before the posterior mean is used for computing acquisition value and level set classification. The adjustment \(u_{dt}\) is an interpolation of the difference \(\textbf{Y}_t-u_p(\textbf{X}_t)\). Specifically, the new posterior mean with adjusted prior is given as:
\begin{align}
    \Tilde{\mu}_t(\textbf{x})=u_p(\textbf{x})+u_{dt}(\textbf{x})+\textbf{k}_t(\textbf{x})^T(\textbf{K}_t+\sigma^2\mathcal{I})^{-1}(\textbf{Y}_t-u_p(\textbf{X}_t)-u_{dt}(\textbf{X}_t))\label{mu_tilde}
\end{align}
where \(u_{dt}=\textbf{k}_t(\textbf{x})^T(\textbf{K}_t+\sigma^2\mathcal{I})^{-1}(\textbf{Y}_t-u_p(\textbf{X}_t))\).  The value of \(u_{dt}(\textbf{x})\) reflects the adjustment of the prior function according to \(\textbf{Y}_t\). When \(\textbf{x}\) is close to points in \(\textbf{X}_t\) (i.e. this region has been sampled), the posterior mean will favor the actual data observations over the prior knowledge. When \(\textbf{x}\) is not close to points in \(\textbf{X}_t\) (i.e. there is still a lot of uncertainty in this region), \(u_{dt}(\textbf{x})\) will be closer to 0 which means \(u_p(\textbf{x})+u_{dt}(\textbf{x})\) becomes similar to \(u_p(\textbf{x})\).

\begin{algorithm2e}
  \caption{Adaptive Prior Level Set Estimation}\label{alg:one}
  \KwIn{prior function \(u_p\), threshold \(h\), budget \(T\).}
  \(D_0\leftarrow\emptyset; t\leftarrow1\)\;
  \While{\(t<T\)}{
    1. Fit the hyperparameters of \(k_t\) with \(\mathcal{GP}(u_p,k(\textbf{x},\textbf{x}'))\) and observations \([\textbf{x}_i,y_i]_{i=1}^t\)\;
    2. Obtain the adjusted prior function: \(u_{p,t}(\textbf{x})=u_p(\textbf{x})+u_{dt}(\textbf{x})\)\;
    3. Obtain the posterior mean function:
    \(\Tilde{\mu}_t(\textbf{x})=u_{p,t}(\textbf{x})+\textbf{k}_t(\textbf{x})^T(\textbf{K}_t+\sigma^2\mathcal{I})^{-1}(\textbf{Y}_t-u_{p,t}(\textbf{X}_t))\)\;
    4. Select next evaluation location \(\textbf{x}_t\) by optimizing a level set acquisition function \(a_t\) over \(\mathcal{X}\) (see examples of \(a_t\) below)\;
    5. Evaluate \(f\) at \(\textbf{x}_t\): \(y_{t}=f(\textbf{x}_t)+\eta_i\)\;
    6. \(D_{t}=D_{t-1}\cup(\textbf{x}_t, y_t)\)\;
    7. \(t=t+1\)
  }
  \KwOut{predicted level set classification\;
  }
\end{algorithm2e}
Details of AP-LSE are in Algorithm \ref{alg:one}. Each iteration first fits kernel hyperparameters using the current observations of \(f\) and uses the fitted kernel to obtain the adjustment function \(u_{dt}\). The posterior mean is computed with the fitted kernel and the adjusted prior function \(u_{p,t}(\textbf{x})=u_p(\textbf{x})+u_{dt}(\textbf{x})\). The location of the next observation is obtained by optimizing an acquisition function \(a_t(\textbf{x})\). Examples of such acquisition functions include:
\begin{itemize}
    \item The Straddle heuristic (\cite{straddle}): \(a_t(\textbf{x})=1.96\sigma_t(\textbf{x})-|\Tilde{\mu}_t(\textbf{x})-h|\)
    \item C2LSE (\cite{c2lse}): \(a_t(\textbf{x})=\frac{\sigma_t(\textbf{x})}{\textup{max}(\epsilon,|\Tilde{\mu}_t(\textbf{x})-h|)}\) with \(\epsilon>0\)
    \item RMILE (\cite{rmile}): \(a_t(\textbf{x})=\textup{max}(\mathbb{E}_{y^+}|I^+|-|I^\epsilon|,\lambda\sigma_t(\textbf{x}))\) for \(\epsilon>0,\lambda>0\), with \(|I^\epsilon|=\sum\limits_{\textbf{x}\in\mathcal{X}}\mathbbm{1}(P(f(\textbf{x})>h-\epsilon)>\delta)\) and \(|I^+|=\sum\limits_{\textbf{x}\in\mathcal{X}}\mathbbm{1}(P^+(f(\textbf{x})>h)>\delta)\) where \(P+\) is calculated based on the GP posterior if we evaluate \(f(\textbf{x})\) and observe value \(y^+\).
\end{itemize}
The algorithm then evaluates \(f\) at the selected location \(\textbf{x}_t\) to obtain the noisy observation \(y_t\), and the new data point \((\textbf{x}_t,y_t)\) joins the existing ones to form the training data \(D_t\). The predicted level set classification can be obtained either at the end of each iteration or when the algorithm terminates by applying classification rules. An example of such rules are:
\begin{itemize}
    \item Predicted superlevel set \(\hat{H}_t=\{\textbf{x}\in\mathcal{X}:\Tilde{\mu}_t(\textbf{x})-\beta\sigma_t(\textbf{x})>h\}\)
    \item Predicted sublevel set \(\hat{L}_t=\{\textbf{x}\in\mathcal{X}:\Tilde{\mu}_t(\textbf{x})+\beta\sigma_t(\textbf{x})<h\}\)
    \item Unclassified set \(\hat{M}_t=\{\textbf{x}\in\mathcal{X}:\Tilde{\mu}_t(\textbf{x})-\beta\sigma_t(\textbf{x})\leq h\leq \Tilde{\mu}_t(\textbf{x})+\beta\sigma_t(\textbf{x})\}\)
\end{itemize}
with confidence parameter \(\beta\).

AP-LSE provides a transfer learning method suitable for most LSE algorithms where \(\Tilde{\mu}_t\) helps us mitigate the use of \(u_p\) when we have sufficient knowledge about \(f\) while still allowing the leverage of the prior information when there is no data about \(f\).

\section{Theoretical Analysis}
Our goal is to show the effectiveness of using \(\Tilde{\mu}_t\) over \(\Bar{\mu}_t\) for active LSE. The analysis first shows that by using \(\Tilde{\mu}_t\), the GP posterior better fits the training data points. We show the upper bounds for the generalization errors of \(\Tilde{\mu}_t\) and \(\Bar{\mu}_t\) and finally a better level set classification confidence using the former.

\subsection{Fitting Error}
One design goal of AP-LSE is for the posterior mean to favor the observations at points close to \(\textbf{X}_t\). This section shows that \(\Tilde{\mu}_t\) is closer to the observed values \(\textbf{Y}_t\) at the observed locations compared to \(\Bar{\mu}_t\), which is demonstrated by a better fitting mean squared error.

\begin{proposition}
    \label{prop:1}
    Let \(\alpha_t(\textbf{x})=\textbf{k}_t(\textbf{x})^T(\textbf{K}_t+\sigma^2\mathcal{I})^{-1}\), \(Z_N=[z_i]_{i=1}^t\) be a vector of constants, and \(\mathcal{H}_k(\mathcal{X})\) be a RKHS corresponding to \(k(\textbf{x}, \textbf{x}')\) on \(\mathcal{X}\). The following inequality holds:
    \begin{align}
        \sum\limits_{i=1}^{t}(z_i-\alpha_t(\textbf{x}_i)Z_t)^2\leq \sum\limits_{i=1}^{t} z_i^2
    \end{align}
    \begin{proof}
        Theorem 3.4 by \cite{kanagawa2018gaussian} stated that \(f^*(\textbf{x})=\alpha_t(\textbf{x})Z_t\) is the solution of the following regularized least-squares problem:
        \begin{align}
            \hat{f}(\textbf{x})=\underset{f\in\mathcal{H}_k}{arg min}\frac{1}{t}\sum\limits_{i=1}^{t}(f(\textbf{x}_i)-y_i)^2+\frac{\sigma^2}{t}||f||_{\mathcal{H}_k}^2\nonumber
        \end{align}
        Therefore, it follows that:
        \begin{align}
            \frac{1}{t}\sum\limits_{i=1}^{t}(f^*(\textbf{x}_i)-y_i)^2+\frac{\sigma^2}{t}||f^*||_{\mathcal{H}_k}^2\leq \frac{1}{t}\sum\limits_{i=1}^{t}(g(\textbf{x}_i)-y_i)^2+\frac{\sigma^2}{t}||g||_{\mathcal{H}_k}^2\textup{ for }g(\textbf{x})=0.
        \end{align}
        This means \(\sum\limits_{i=1}^{t}(z_i-\alpha_t(\textbf{x}_i)Z_t)^2+\sigma^2||\alpha_t(\textbf{x})Z_t||_{\mathcal{H}_k}^2 \leq \sum\limits_{i=1}^{t} z_i^2\), or the inequality holds.
    \end{proof}
\end{proposition}
The following result states that AP-LSE fits the training data points better than vanilla transfer learning.
\begin{theorem}
    Let \(\mathcal{D}_t=[\textbf{x}_i,y_i]_{i=1}^t\) be the observations of \(f\). The fitting error of \(\Tilde{\mu}_t(\textbf{x})\) is upper bounded by that of \(\Bar{\mu}_t(\textbf{x})\), which is bounded by the difference between \(u_p\) and \(f\). That is:
    \begin{align}
        \sum\limits_{i=1}^{t}(\Tilde{\mu}_t(\textbf{x}_i)-y_i)^2\leq \sum\limits_{i=1}^{t}(\Bar{\mu}_t(\textbf{x}_i)-y_i)^2\leq \sum\limits_{i=1}^{t}(u_p(\textbf{x}_i)-y_i)^2 \label{better_training_error}
    \end{align}
    \begin{proof}
        We have that:
        \begin{align}
            \Tilde{\mu}_t(\textbf{x}_i)-y_i&=u_p(\textbf{x}_i)+u_{dt}(\textbf{x}_i)+\alpha_t(\textbf{x}_i)(\textbf{Y}_t-u_p(\textbf{X}_t)-u_{dt}(\textbf{X}_t))-y_i\nonumber\\
            &=u_p(\textbf{x}_i)+\alpha_t(\textbf{x}_i)(\textbf{Y}_t-u_p(\textbf{X}_t))-y_i+\alpha_t(\textbf{x}_i)(\textbf{Y}_t-u_p(\textbf{X}_t)-\alpha_t(\textbf{X}_t)(\textbf{Y}_t-u_p(\textbf{X}_t)))\nonumber\\
            &=\Bar{\mu}_t(\textbf{x}_i)-y_i+\alpha_t(\textbf{x}_i)(\textbf{Y}_t-\Bar{\mu}_t(\textbf{X}_t))\nonumber\\
            &=u(\textbf{x}_i)-\alpha_t(\textbf{x}_i)u(\textbf{X}_t),\nonumber
        \end{align}
        for \(1\leq i\leq t\) with \(u(\textbf{x})=\Bar{\mu}_t(\textbf{x})-y\), \(y\) being the observation of \(f\) at \(\textbf{x}\). Using Proposition \ref{prop:1}, we can establish the left inequality in (\ref{better_training_error}) as follows:
        \begin{align}
            \sum\limits_{i=1}^{t}(u(\textbf{x}_i)-\alpha_t(\textbf{x}_i)u(\textbf{X}_t))^2&\leq \sum\limits_{i=1}^{t} u(\textbf{x}_i)^2=\sum\limits_{i=1}^{t}(\Bar{\mu}_t(\textbf{x}_i)-y_i)^2\nonumber
        \end{align}
        Let \(g(\textbf{x})=u_p(\textbf{x})-y\). We then have 
        \begin{align}
            u(\textbf{x}_i)&=u_p(\textbf{x}_i)-y+\alpha_t(\textbf{x}_i)(\textbf{Y}_t-u_p(\textbf{X}_t))=g(\textbf{x})-\alpha_t(\textbf{x})g(\textbf{X}_t)
        \end{align}
        Using Proposition \ref{prop:1}, we can establish the right inequality in (\ref{better_training_error}):
        \begin{align}
            \sum\limits_{i=1}^{t}u(\textbf{x}_i)^2 &=\sum\limits_{i=1}^{t}(g(\textbf{x}_i)-\alpha_t(\textbf{x})g(\textbf{X}_t))^2\leq \sum\limits_{i=1}^{t} g(\textbf{x}_i)^2\nonumber
        \end{align}
        Both inequalities in (\ref{better_training_error}) hold as a result.
    \end{proof}
\end{theorem}
\vspace{-30pt}
\subsection{Generalization Error}
In addition to the fitting error at observed locations, it is more important that a model generalizes well at other locations. Here we give bounds to the generalization errors of \(\Tilde{\mu}_t\) and \(\Bar{\mu}_t\). Let us denote \(\Bar{e}_t(\textbf{x})=\Bar{\mu}_{t-1}(\textbf{x})-f(\textbf{x})\) and \(\Tilde{e}_t(\textbf{x})=\Tilde{\mu}_{t-1}(\textbf{x})-f(\textbf{x})\). 

\begin{lemma}\label{lma:gen_bound}
    Let \(\rho_t=\underset{\textbf{x}\in\mathcal{X}}{\textup{sup}}\underset{\textbf{x}_i\in\textbf{X}_t}{\textup{inf}}\norm{\textbf{x}-\textbf{x}_i}_2\). Let \(y\) be the observation of \(f\) at \(\textbf{x}\). Assume that:
    \begin{enumerate}
        \item \(f\in\mathcal{H}_k(\mathcal{X})\),
        \item \(\mathcal{H}_k(\mathcal{X})\) is isomorphic to the Sobolev space \(\mathcal{H}^{\tau}(\mathcal{X})\) for some \(\tau=n+r\), with \(n\in\mathbb{N},n>d/2\) and \(0<r<1\),
        \item \(u_p,f\in\mathcal{H}^{\tau}(\mathcal{X})\),
        \item The Gaussian Process uses radial basis function (RBF) kernel.
    \end{enumerate}    
    Then there exist a constant C independent of \(f\), \(u_p\), and \(t\), we have for \(0\leq\xi\leq\tau\):
    \begin{align}
        \norm{\Bar{e}_t}_{\mathcal{H}^\xi(\mathcal{X})}&\leq C\rho^{\tau-\xi}_t\norm{u_p-f}_{\mathcal{H}^{\tau}(\mathcal{X})}+\norm{\alpha_t\textbf{E}_t}_{\mathcal{H}^\xi(\mathcal{X})},\label{bar_bound}\\
        \text{and }\norm{\Tilde{e}_t}_{\mathcal{H}^\xi(\mathcal{X})}&\leq C\rho^{\tau-\xi}_t\norm{\Bar{e}_t}_{\mathcal{H}^{\tau}(\mathcal{X})}+\norm{\alpha_t\textbf{E}_t}_{\mathcal{H}^\xi(\mathcal{X})},\label{tilde_bound}
    \end{align}
    \begin{proof}
        We have that:
        \begin{align}
            \norm{\Bar{\mu}_{t-1}-f}_{\mathcal{H}^\xi(\mathcal{X})}&=\norm{u_p-f-\alpha_t(u_p(\textbf{X}_t)-\textbf{Y}_t)}_{\mathcal{H}^\xi(\mathcal{X})}\nonumber\\
            &\leq \norm{u_p-f-\alpha_t(u_p(\textbf{X}_t)-f(\textbf{X}_t))}_{\mathcal{H}^\xi(\mathcal{X})}+\norm{\alpha_t\textbf{E}_t}_{\mathcal{H}^\xi(\mathcal{X})}\label{base}
        \end{align}
        Given the assumptions, we have the following using Lemma 4.1 in \cite{narcowich1}:
        \begin{align}
            \norm{u_p-f-\alpha_t(u_p(\textbf{X}_t)-f(\textbf{X}_t))}_{\mathcal{H}^\xi(\mathcal{X})}\leq C\rho^{\tau-\xi}_t\norm{u_p-f}_{\mathcal{H}^{\tau}(\mathcal{X})}\label{inq:11}
        \end{align}
        Combining Ineq.\ref{inq:11} with Ineq.\ref{base} establishes Ineq.\ref{bar_bound}.
        Ineq.\ref{tilde_bound} can be established similarly.
    \end{proof}
\end{lemma}
\vspace{-15pt}
The first assumption is a general one on \(f\) being from the native space \(\mathcal{H}_k(\mathcal{X})\), similar to \cite{srinivas}. The second and third assumptions mean that both the prior function \(u_p\) and the true underlying function \(f\) have similar smoothness \(\tau\) in the Sobolev space \(\mathcal{H}^{\tau}(\mathcal{X})\) which is equal to the native space \(\mathcal{H}_k(\mathcal{X})\). The last assumption limits the following theoretical results to only LSE algorithms that employ an RBF kernel for the GP, but our experiments show AP-LSE is still effective with other kernels.

\begin{lemma}
    \label{lma:better_convergence}
    When applied to an LSE algorithm that eventually evaluates every point in an arbitrarily dense finite grid, the generalization errors of both vanilla transfer learning and AP-LSE converge to zero, that is
    \begin{align}
        \underset{t\rightarrow\infty}{lim}|\Bar{e}_t(\textbf{x})|=0\textup{ and }\underset{t\rightarrow\infty}{lim}|\Tilde{e}_t(\textbf{x})|=0.
    \end{align}
    In addition, let \(\Bar{r}\) and \(\Tilde{r}\) be the convergence rates for the upper bounds of \(\{\norm{\Bar{e}_t}_{\mathcal{H}^\xi(\mathcal{X})}\}_{t=1}^{\infty}\) and \(\{\norm{\Tilde{e}_t}_{\mathcal{H}^\xi(\mathcal{X})}\}_{t=1}^{\infty}\) in Lemma \ref{lma:gen_bound} respectively. Then \(\Bar{r}<\Tilde{r}\).
    \begin{proof}
        Examples of LSE algorithms that eventually evaluates every point in an arbitrarily dense finite grid include C2LSE and RMILE (as shown in the proof of Theorem 6 by \cite{c2lse} and Lemma 4 by \cite{rmile}, respectively). This means that \(\textbf{X}_t\rightarrow\mathcal{X}\) as \(t\rightarrow\infty\) for a continuous domain \(\mathcal{X}\), or equivalently \(\underset{t\rightarrow\infty}{lim}\rho_t=0\). The first consequence is that \(\underset{t\rightarrow\infty}{lim}C\rho^{\tau-\xi}_t\norm{u_p-f}_{\mathcal{H}^{\tau}(\mathcal{X}_{\rho,t})}=0\).

        Secondly, \cite{10.5555/3305381.3305469} showed that:
        \begin{align}
            |\alpha_t(\textbf{x})\textbf{E}_t|\leq\sigma^{-1/2}\sigma_{t-1}(\textbf{x})\sqrt{\textbf{E}_t^TK_t(K_t+\sigma^2\mathcal{I})^{-1}\textbf{E}_t}\nonumber
        \end{align}
        If \(\textbf{X}_t\rightarrow\mathcal{X}\) as \(t\rightarrow\infty\), we have \(\underset{t\rightarrow\infty}{lim}|\sigma_t(\textbf{x})|=0\), so \(\underset{t\rightarrow\infty}{lim}|\alpha_t(\textbf{x})\textbf{E}_t|=0\) and eventually, \(\underset{t\rightarrow\infty}{lim}\norm{\alpha_t\textbf{E}_t}_{\mathcal{H}^\xi(\mathcal{X})}=0\). As a result, we have that \(\underset{t\rightarrow\infty}{lim}\norm{\Bar{\mu}_t-f}_{\mathcal{H}^\xi(\mathcal{X}_{\rho,t})}=0\), and consequently, \(\underset{t\rightarrow\infty}{lim}|\Bar{e}_t(\textbf{x})|=0\). The convergence of AP-LSE's error can be established similarly.
    
        Regarding convergence rates, we have that:
        \begin{align}
            \Bar{r}&=\underset{t\rightarrow\infty}{lim}\frac{C\rho^{\tau-\xi}_{t+1}\norm{u_p-f}_{\mathcal{H}^{\tau}(\mathcal{X})}+\norm{\alpha_{t+1}\textbf{E}_{t+1}}_{\mathcal{H}^\xi(\mathcal{X})}}{C\rho^{\tau-\xi}_t\norm{u_p-f}_{\mathcal{H}^{\tau}(\mathcal{X})}+\norm{\alpha_t\textbf{E}_t}_{\mathcal{H}^\xi(\mathcal{X})}},\label{first_rate}\\
            \textup{and }\Tilde{r}&=\underset{t\rightarrow\infty}{lim}\frac{C\rho^{\tau-\xi}_{t+1}\norm{\Bar{e}_{t+1}}_{\mathcal{H}^{\tau}(\mathcal{X})}+\norm{\alpha_{t+1}\textbf{E}_{t+1}}_{\mathcal{H}^\xi(\mathcal{X})}}{C\rho^{\tau-\xi}_t\norm{\Bar{e}_t}_{\mathcal{H}^{\tau}(\mathcal{X})}+\norm{\alpha_t\textbf{E}_t}_{\mathcal{H}^\xi(\mathcal{X})}} \label{second_rate}
        \end{align}
        In Eq.\ref{first_rate}, it can be seen that only the convergences of \(\rho_t\) and \(\norm{\alpha_t\textbf{E}_t}_{\mathcal{H}^\xi(\mathcal{X})}\) contribute to \(\Bar{r}\) while \(\norm{u_p-f}_{\mathcal{H}^{\tau}(\mathcal{X})}\) remains fixed. Compared to Eq.\ref{first_rate}, Eq.\ref{second_rate} replaces \(\norm{u_p-f}_{\mathcal{H}^{\tau}(\mathcal{X})}\) with \(\norm{\Bar{e}_{t+1}}_{\mathcal{H}^{\tau}(\mathcal{X})}\) which converges to 0. Therefore, we can conclude that \(\Bar{r}<\Tilde{r}\).
    \end{proof}
\end{lemma}
\vspace{-30pt}
\subsection{Level Set Classification}
\begin{theorem}\label{thm:better_lse_convergence}
    Given a continuous LSE problem, assume that C2LSE with confidence parameter \(\beta\) reaches an \(\epsilon\)-accurate solution at iteration \(T\) for any point \(\textbf{x}\in\mathcal{X}\) with probabilities \(\Bar{\delta}\) using vanilla transfer learning. Then C2LSE using AP-LSE also reaches an \(\epsilon\)-accurate solution at the same iteration for any point \(\textbf{x}\in\mathcal{X}\) with probabilities \(\Tilde{\delta}\) where \(\Tilde{\delta}>\Bar{\delta}\).
    \begin{proof}
        According to Theorem 9 by \cite{c2lse}, C2LSE with vanilla transfer learning reaches an \(\epsilon\)-accurate solution for any point \(\textbf{x}\in\mathcal{X}\) with probability \(\Bar{\delta}\) if:
        \begin{align}
            T\geq\frac{8\beta\gamma_T}{\epsilon^2log(1+\sigma^{-2})}\label{c2lse_convergence_rate}
        \end{align}
        assuming \(\textup{Pr}(|\Bar{e}_t(\textbf{x})|\leq\beta\sigma_{t-1}(\textbf{x}))=\Bar{\delta},\forall \textbf{x}\in \mathcal{X}, \forall t\geq1\) where \(\epsilon>0\), and \(\gamma_T\) is the maximum information gain about \(f\) over the set of all possible \(T\) noisy observations \(\mathcal{D}_T\).

        Lemma \ref{lma:better_convergence} indicates that the upper bound of \(\{\norm{\Tilde{e}_t}_{H^\xi(\mathcal{X})}\}_{i=1}^{\mathcal{N}^+}\) converges to 0 faster than the upper bound of \(\{\norm{\Bar{e}_t}_{H^\xi(\mathcal{X})}\}_{i=1}^{\mathcal{N}^+}\). This implies that for each \(\textbf{x}\), there exist \(\Delta_\textbf{x}>0\) such that \(|\Bar{e}_t(\textbf{x})|=|\Tilde{e}_t(\textbf{x})|+\Delta_\textbf{x}\). We then have that:
        \begin{align}
            \textup{Pr}(|\Tilde{e}_t(\textbf{x})|\leq\beta\sigma_{t-1}(\textbf{x}))&=\textup{Pr}(|\Bar{e}_t(\textbf{x})|-\Delta_\textbf{x}\leq\beta\sigma_{t-1}(\textbf{x}))\nonumber\\
            &<\textup{Pr}(|\Bar{e}_t(\textbf{x})|\leq\beta\sigma_{t-1}(\textbf{x}))=\Bar{\delta},\forall \textbf{x}\in \mathcal{X}, \forall t\geq1
        \end{align}
        Assuming \(\textup{Pr}(|\Tilde{e}_t(\textbf{x})|\leq\beta\sigma_{t-1}(\textbf{x}))=\Tilde{\delta},\forall \textbf{x}\in \mathcal{X}, \forall t\geq1\), we then have \(\Bar{\delta}<\Tilde{\delta}\). The remaining steps to obtain the corresponding convergence rate and level set classification accuracy are similar to both vanilla transfer learning and AP-LSE and are independent of \(\Tilde{\delta}\) and \(\Bar{\delta}\). As such, C2LSE using AP-LSE also reaches an \(\epsilon\)-accurate solution at the same iteration for any point \(\textbf{x}\in\mathcal{X}\) with probabilities \(\Tilde{\delta}\).
    \end{proof}
\end{theorem}
\vspace{-15pt}
Theorem \ref{thm:better_lse_convergence} indirectly shows a better sample efficiency for AP-LSE. It proves that by using AP-LSE, the probability that a point is classified correctly will be higher than by using vanilla transfer learning at the same iteration. Therefore, over the whole input space \(\mathcal{X}\), more points will be classified correctly by using AP-LSE with the same evaluation budget. This is equivalent to AP-LSE achieving the same classification accuracy with vanilla transfer learning while using a lower evaluation budget. While the result is only proven for C2LSE, our experiments in the following section show that AP-LSE also improves the sample efficiency for other LSE algorithms.

\section{Numerical Experiments}\label{Section5}
\subsection{Experiment Setup}
We will study the performance of AP-LSE on transfer learning for multiple LSE tasks by using it in combination with existing LSE algorithms, namely STR (the Straddle heuristic by \cite{straddle}), RMILE (\cite{rmile}), and C2LSE (\cite{c2lse}). STR is similar in spirit to the GP-UCB algorithm (\cite{srinivas}), while C2LSE is a confidence-based method for continuous LSE problems. RMILE is another approach for LSE influenced by the expected improvement algorithm (\cite{Brochu2010ATO}) and operates on discrete domains. These three algorithms comprise a comprehensive test for the effectiveness of transfer learning across different styles of LSE algorithms.

We compare AP-LSE with running LSE from scratch and vanilla transfer learning to demonstrate the sample efficiency of each method across each LSE algorithm. For a fair comparison, we start all methods without initialized data points so that the classification performance of all three methods starts similarly. Though our setting does not consider the availability of previous observations of a related LSE task (i.e. source data), we include Diff-GP, which utilizes such data, to compare between directly using the source data like Diff-GP and using a surrogate model fitted on source data like AP-LSE. Diff-GP first models the difference between the target function's observations and the posterior mean fitted on the source data and uses that difference to adjust the source data before incorporating them into the posterior GP for the target function.

We report the average F1-score among 30 runs in each experiment. This is calculated based on the true level set labels of a fixed finite grid for each problem. Hyperparameters are selected based on the recommended values in the original works. We use a Mat\'ern kernel with hyperparameters fitted by maximum likelihood in the section.

\noindent\textbf{Synthetic Functions}

\noindent We use the following synthetic functions:
\begin{itemize}
    \item Bird: \(f(x_1, x_2)=sin(x_1)e^{(1+\kappa-cos(x_2))^2}+cos(x_2)e^{(1-sin(x_1))^2}+(x_1-x_2)^2\)    
    \item Multi-circle 3D (MC3D): \(f(x_1,x_2,x_3)=e^{(sin(x_1+\kappa))^2\times(sin(x_2+\kappa))^2\times(sin(x_3+\kappa))^2}\)   
    \item Mishra03: \(f(x_1,x_2)=\sqrt{|cos(\sqrt{x_1^2+x_2^2}+\kappa)|}+0.01(x_1+x_2)\)
\end{itemize}
The goal for each function is to find its superlevel or sublevel set given a function evaluation budget \(T\). We discretize their input space into a uniform grid to evaluate the quality of level set classification. The parameter \(\kappa\) in these functions is to differentiate the prior and the target functions. The prior function is obtained by first setting \(\kappa=0\) and evaluating the function at \(T\) random locations. We fit a GP on these observations and use its posterior mean as the prior function for AP-LSE and vanilla transfer learning. Diff-GP directly uses these observations as its source data to simulate a similar amount of prior information between Diff-GP and the transfer learning methods (i.e. AP-LSE and vanilla transfer learning). The target function is obtained by using a different value for \(\kappa\). The observation noise level is set at 0.1 for all functions. Details about each function are as follows:
\begin{table}[htbp!]
    \centering
    \begin{tabular}{|c|c|c|c|c|c|} 
        \hline
        & \textbf{Input space} & \textbf{Goal} & \textbf{Evaluation grid} & \textbf{Budget \(T\)} & \textbf{Target} \(\kappa\) \\
        \hline
        Bird & \(\textbf{x}\in[-6,6]^2\) & \(f(\textbf{x})<4\) & 100\(\times\)100 & 150 & 0.4\\
        \hline
        MC3D & \(\textbf{x}\in[0,6]^3\) & \(f(\textbf{x})>1.6\) & 20\(\times\)20\(\times\)20 & 400 &0.3\\
        \hline
        Mishra03 & \(\textbf{x}\in[-5,5]^2\) & \(f(\textbf{x})<0.7\) & 100\(\times\)100 & 250 &0.4\\
        \hline
    \end{tabular}
    \caption{Details about synthetic functions}
    \label{tab:synthetic}
\end{table}

\noindent\textbf{Real-world Data}

\noindent We also evaluate transfer learning performance on three real-world LSE tasks. Each task includes a source dataset and a target dataset. Their details are as follows:
\begin{itemize}
    \item Defective (Defective Area Identification (\cite{hozumi2023adaptive})):  From an active LSE perspective, the defective areas on a material surface can be estimated using only a small number of informative measurements instead of measuring all grid points which can be both slow and costly. We use the dataset provided by \cite{hozumi2023adaptive} about red-zone identification for solar cell ingots in which the red-zone is the defective area with low carrier lifetime values. The dataset includes carrier lifetime values for two ingots (one chosen as the source dataset and the other as the target dataset), both in the form of a 161\(\times\)121 uniform grid where the value is recorded for each grid point. The target is to find the sublevel set that has carrier lifetime values under 80.
    \item Network (Network Latency): Inspired by \cite{activelse}, we create a network latency dataset that measures the round-trip time around the world. We ping 223 servers on Dec 12th, 2023 (source dataset) and 225 servers on Dec 13th, 2023 (target dataset) to record their round-trip times. The input space includes each server's longitude and latitude coordinates. The target is to find the sublevel set that has round-trip times under 200ms indicating good internet quality.
    \item AA (Algorithmic Assurance): This task attempts to find scenarios in which a machine learning model's performance is at least above a certain threshold. Specifically, we train a convolutional neural network on MNIST and find its accuracy on different transformation scenarios for the test set. The input space includes continuous values of the transformations applied (i.e. rotation, scaling, and shear), and the output is the classification error on the transformed test set. Details about this task can be found in the work of \cite{c2lse}. We construct the source dataset by training on 75\% of MNIST images and the target dataset by training on all images. Each dataset includes a grid of 47,916 data points as 36\(\times\)11\(\times\)11\(\times\)11 uniform grid. The target is to find the sublevel set that has classification error on the test set of MNIST under 4\%.
\end{itemize}
Similar to the synthetic functions, a number of random data points from the source dataset, which is equal to the evaluation budget (50 for Defective, 250 for Network, and 300 for AA), are used to obtain a posterior mean serving as the prior function for AP-LSE and vanilla transfer learning and are used directly as source data for Diff-GP. The target function is obtained by selecting a random subset of data points in the target dataset (2\% for Defective and AA, and 100\% for Network) and then fitting a GP posterior with which the posterior mean becomes the target function.

\begin{table}[htbp!]
    \centering
    \begin{tabular}{c}
        \vspace{0.3cm}
        \begin{minipage}{\dimexpr\linewidth-2\fboxsep\relax}
            \centering
            \includegraphics[width=\textwidth]{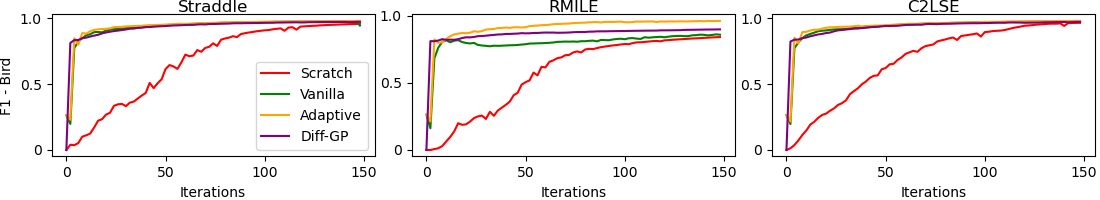}
        \end{minipage} \\
        \vspace{0.3cm}
        \begin{minipage}{\dimexpr\linewidth-2\fboxsep\relax}
            \centering
            \includegraphics[width=\textwidth]{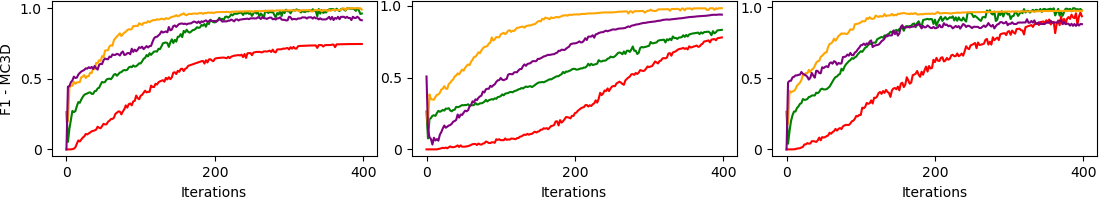}\par
        \end{minipage}\\
        \begin{minipage}{\dimexpr\linewidth-2\fboxsep\relax}
            \centering
            \includegraphics[width=\textwidth]{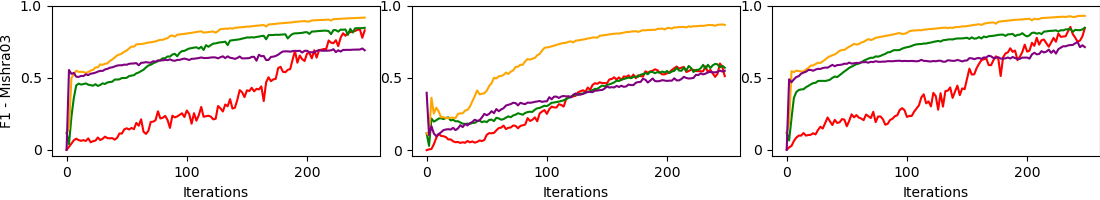}\par
        \end{minipage}\\
    \end{tabular}
    \caption*{Figure 2: Transfer learning for active LSE on synthetic functions}
\end{table}
\subsection{Main Results}
Figure 2 shows that AP-LSE consistently achieves better classification performances for all three functions regardless of the LSE algorithms employed. The performance gaps become smaller in the later iterations, which is expected when the transferred knowledge's effect decreases with more data observed. Similar patterns are observed in Figure 3 with better performances for AP-LSE compared to the other two transfer learning methods. For both synthetic functions and real-world problems, AP-LSE demonstrates superior classification accuracy from the beginning compared to vanilla transfer learning. This matches the goal of active LSE where high classification accuracy using few data points is desirable.

\begin{table}
    \centering
    \begin{tabular}{c}
        \vspace{0.3cm}
        \begin{minipage}{\dimexpr\linewidth-2\fboxsep\relax}
            \centering
            \includegraphics[width=\textwidth]{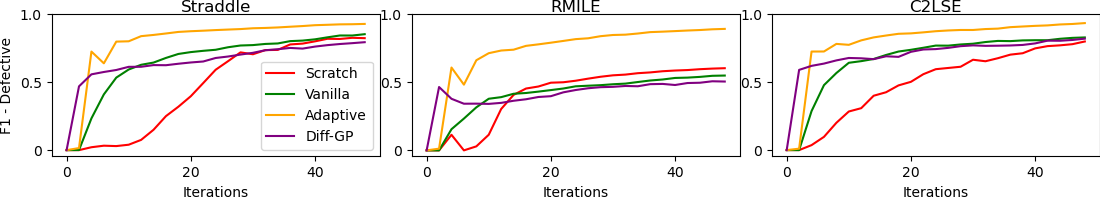}
        \end{minipage} \\ 
        \vspace{0.3cm}
        \begin{minipage}{\dimexpr\linewidth-2\fboxsep\relax}
            \centering
            \includegraphics[width=\textwidth]{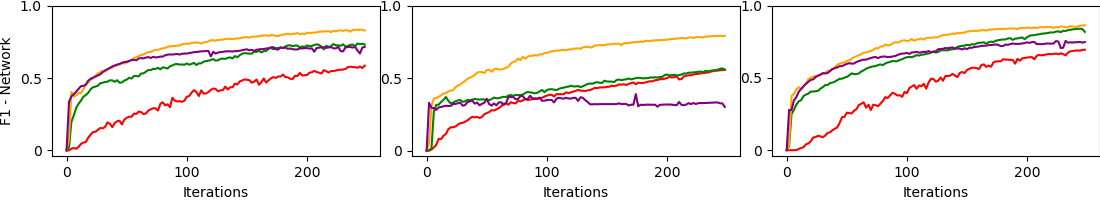}\par
        \end{minipage}\\
        \begin{minipage}{\dimexpr\linewidth-2\fboxsep\relax}
            \centering
            \includegraphics[width=\textwidth]{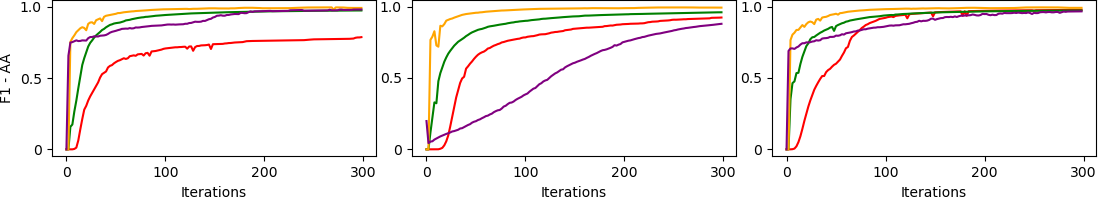}\par
        \end{minipage}\\
    \end{tabular}
    \caption*{Figure 3: Transfer learning for active LSE on real-world problems}
\end{table} 

AP-LSE outperforms Diff-GP on all datasets in the long term, which indicates using an adaptive surrogate model as prior for active LSE is more effective than directly involving source data like Diff-GP even though Diff-GP takes into account the difference between the source task and the target task. We attribute this performance difference to the struggle of Diff-GP to capture the source-target difference. As noted by the authors of Diff-GP, its performance depends on the simplicity of learning this difference and the proximity of the source data to the target evaluation locations, so it struggles when the difference is too complex to model (as demonstrated in the next section). Our method does not have those dependencies. We do not use source data and seamlessly incorporate the observed difference into our posterior rather than modeling it separately. As such, AP-LSE is not affected by the difficulty in modeling the difference between the prior and the target function.

\subsection{Varying Prior Level Set Similarity}
The main design goal of AP-LSE is to have robust classification performance for an LSE algorithm regardless of the prior's relevance to the target function. We now evaluates the methods on different levels of level set similarity between the prior and the target function measured by the percentage of grid points with the same level set label between the prior and the target function. We perform this experiment on Mishra03 where the prior function remains as described in Section 5.1 (\(\kappa=0\)), and the target function varies by varying \(\kappa\).
\begin{table}[htbp!]
    \centering
    \begin{tabular}{|c|c|c|c|c|c|c|} 
         \hline
         \multicolumn{2}{|c|}{\(\kappa\)} & 0.2 & 0.4 & 0.6 & 0.8 & 1.0\\
         \hline
         \multicolumn{2}{|c|}{Level set similarity} & 88\% & 77\% & 67\% & 56\% & 47\% \\
         \hline
         \multirow{4}{*}{Straddle}&Scratch & 243 & 229 & 205 & \textbf{158} & \textbf{173}\\ 
         \cline{2-7}
         &Vanilla & 88 & 160 & 217 & 215 & 207\\
         \cline{2-7}
         & DiffGP & 59 & NA & NA & NA & NA \\
         \cline{2-7}
         &AP-LSE & \textbf{24} & \textbf{84} & \textbf{150} & 168 & \textbf{173}\\
         \hline
         \multirow{4}{*}{C2LSE}&Scratch & 298 & 246 & 255 & 183 & 177 \\
         \cline{2-7}
         &Vanilla & 78 & 189 & 250 & 212 & 198\\
         \cline{2-7}
         & DiffGP & 56 & NA & NA & NA & NA \\
         \cline{2-7}
         &AP-LSE & \textbf{20} & \textbf{85} & \textbf{154} & \textbf{170} & \textbf{165}\\
         \hline
    \end{tabular}
    \caption{\#Iterations to reach 80\% F1-score on Mishra03 with different similarity levels. NA means the algorithm cannot reach the target F1-score within the evaluation budget.}
    \label{tab:varying}
\end{table}

Table \ref{tab:varying} shows the average number of iterations needed to reach 80\% F1-score on Mishra03. On mostly all similarity levels, AP-LSE requires significantly fewer iterations than other methods, even at lower similarity. Diff-GP only works well when the difference is small as expected. It is also noteworthy that AP-LSE may not have an advantage over running from scratch if the prior function is largely different from the target function.
\section{Conclusion}
We proposed AP-LSE, a transfer learning method for active LSE that robustly incorporates the prior function such that an LSE algorithm can utilize the prior information while still being safeguarded from discrepancies between the prior and the target black-box function. AP-LSE locally adjusts the prior function according to the observed discrepancies to provide more accurate function estimation in level set classification. A detailed theoretical analysis shows that our method is effective in improving function estimation and consequently, in improving level set classification by increasing the classification confidence for the same accuracy. Our empirical results confirm the superior performance of AP-LSE compared to vanilla transfer learning on various active LSE tasks using different LSE algorithms.

\acks{This research was partially supported by the Australian Government through the Australian Research Council's Discovery Projects funding scheme (project DP210102798). The views expressed herein are those of the authors and are not necessarily those of the Australian Government or Australian Research Council.}
\bibliography{acml24}
\newpage
\appendix
\section{Results with other kernels}\label{apd:first}
We replicate the experiments in Section 5.2 with the radial basis function (RBF) and inverse multi-quadric (IMQ) kernels. 
\begin{table}[htbp!]
    \centering
    \begin{tabular}{c}
        \begin{minipage}{\dimexpr\linewidth-2\fboxsep\relax}
            \centering
            \includegraphics[width=0.95\textwidth]{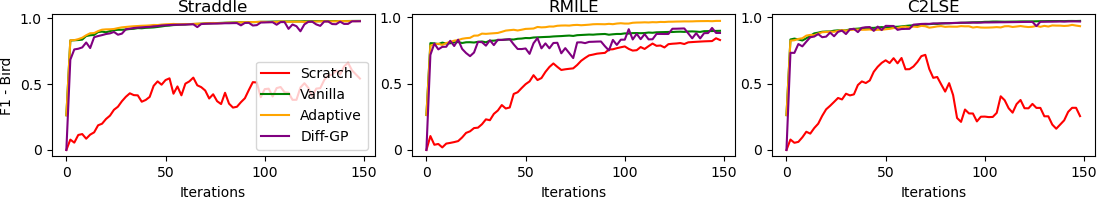}
        \end{minipage} \\
        \begin{minipage}{\dimexpr\linewidth-2\fboxsep\relax}
            \centering
            \includegraphics[width=0.95\textwidth]{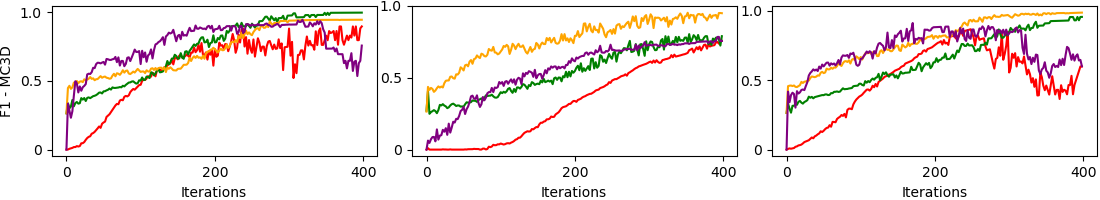}\par
        \end{minipage}\\
        \begin{minipage}{\dimexpr\linewidth-2\fboxsep\relax}
            \centering
            \includegraphics[width=0.95\textwidth]{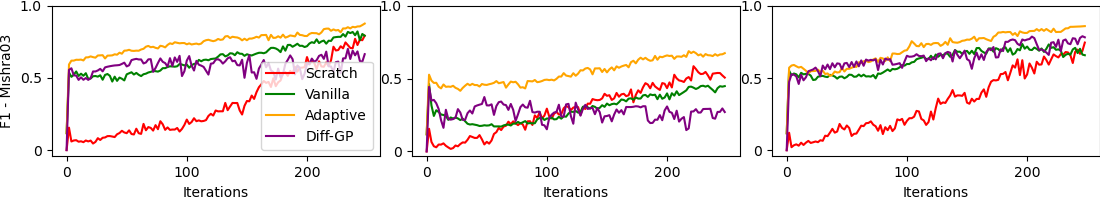}\par
        \end{minipage}\\
        \begin{minipage}{\dimexpr\linewidth-2\fboxsep\relax}
            \centering
            \includegraphics[width=0.95\textwidth]{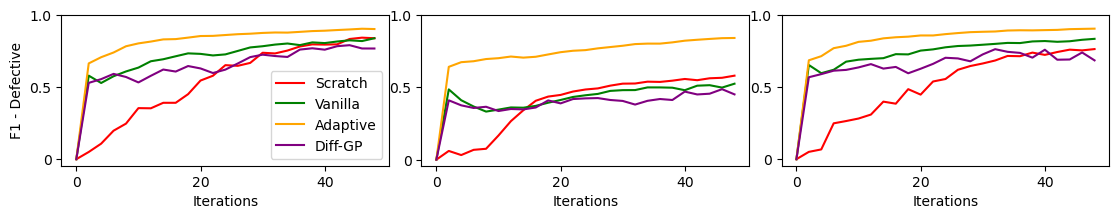}
        \end{minipage} \\ 
        \begin{minipage}{\dimexpr\linewidth-2\fboxsep\relax}
            \centering
            \includegraphics[width=0.95\textwidth]{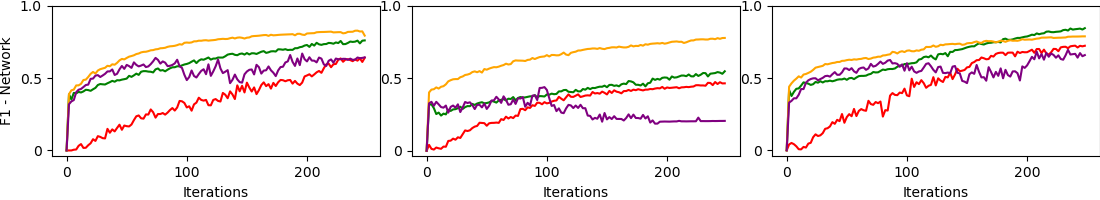}\par
        \end{minipage}\\
        \begin{minipage}{\dimexpr\linewidth-2\fboxsep\relax}
            \centering
            \includegraphics[width=0.95\textwidth]{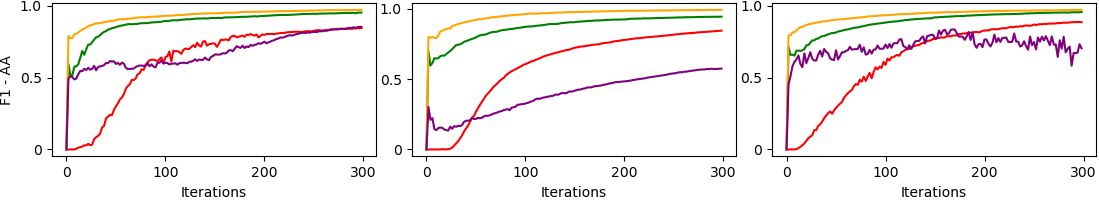}\par
        \end{minipage}\\
    \end{tabular}
    \caption*{Figure 4: Transfer learning for active LSE with RBF kernel}
\end{table}
\newpage
\begin{table}[htbp!]
    \centering
    \begin{tabular}{c}
        \begin{minipage}{\dimexpr\linewidth-2\fboxsep\relax}
            \centering
            \includegraphics[width=0.95\textwidth]{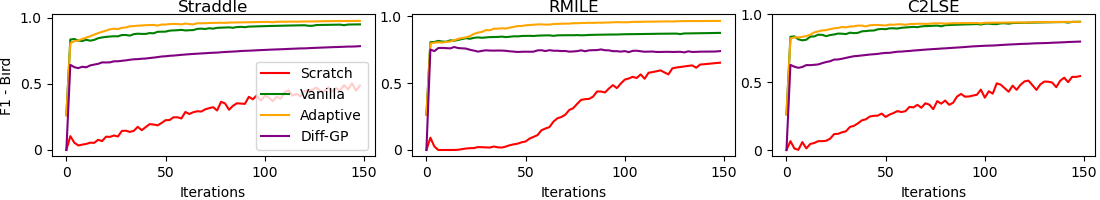}
        \end{minipage} \\
        \begin{minipage}{\dimexpr\linewidth-2\fboxsep\relax}
            \centering
            \includegraphics[width=0.95\textwidth]{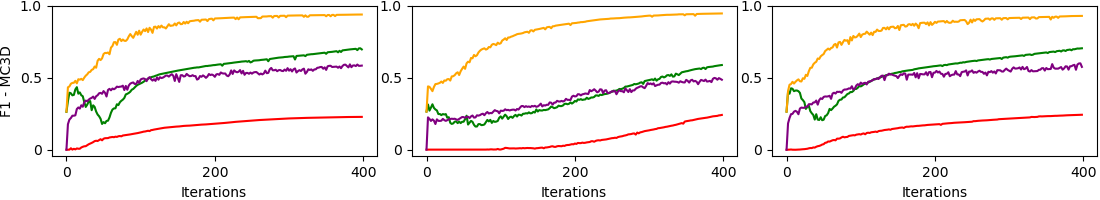}\par
        \end{minipage}\\
        \begin{minipage}{\dimexpr\linewidth-2\fboxsep\relax}
            \centering
            \includegraphics[width=0.95\textwidth]{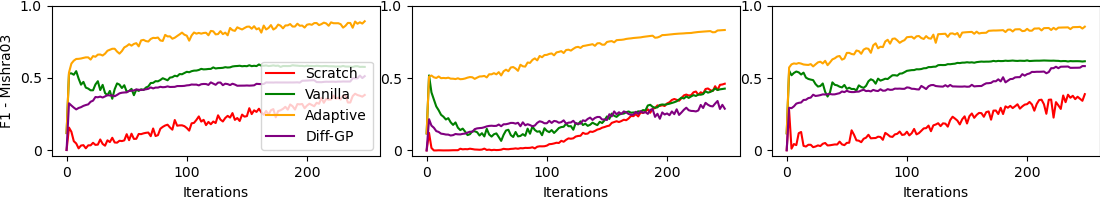}\par
        \end{minipage}\\
        \begin{minipage}{\dimexpr\linewidth-2\fboxsep\relax}
            \centering
            \includegraphics[width=0.95\textwidth]{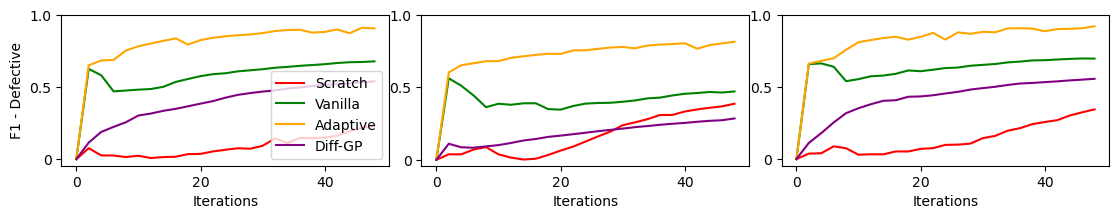}
        \end{minipage} \\ 
        \begin{minipage}{\dimexpr\linewidth-2\fboxsep\relax}
            \centering
            \includegraphics[width=0.95\textwidth]{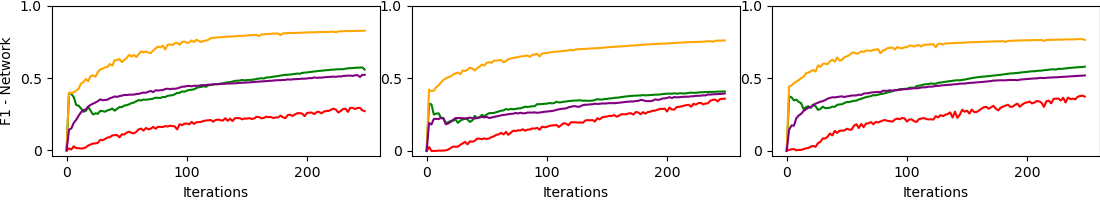}\par
        \end{minipage}\\
        \begin{minipage}{\dimexpr\linewidth-2\fboxsep\relax}
            \centering
            \includegraphics[width=0.95\textwidth]{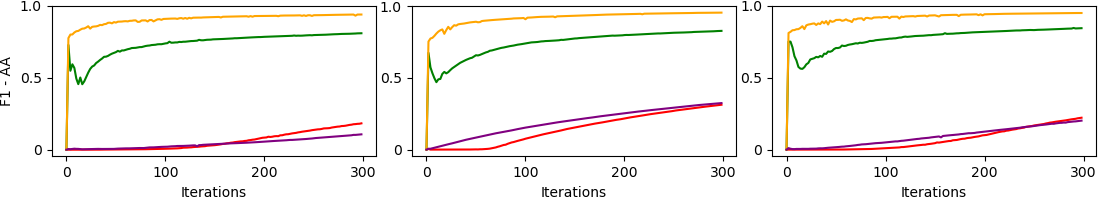}\par
        \end{minipage}\\
    \end{tabular}
    \caption*{Figure 5: Transfer learning for active LSE with IMQ kernel}
\end{table}
\end{document}